\documentclass[letterpaper]{article} 
\usepackage{aaai2026}  
\usepackage{times}  
\usepackage{helvet}  
\usepackage{courier}  
\usepackage[hyphens]{url}  
\usepackage{graphicx} 
\usepackage{natbib}  
\usepackage{caption} 
\usepackage{algorithm}
\usepackage{algorithmic}

\usepackage{newfloat}
\usepackage{listings}
\DeclareCaptionStyle{ruled}{labelfont=normalfont,labelsep=colon,strut=off} 
\floatstyle{ruled}
\newfloat{listing}{tb}{lst}{}
\floatname{listing}{Listing}
\title{Dexterous Manipulation Transfer via Progressive Kinematic-Dynamic Alignment}
\author{
    Wenbin Bai\textsuperscript{\rm 1},
    Qiyu Chen\textsuperscript{\rm 1},
    Xiangbo Lin\textsuperscript{\rm 1}\thanks{Corresponding author.},
    Jianwen Li\textsuperscript{\rm 2},\\
    Quancheng Li\textsuperscript{\rm 1},
    Hejiang Pan\textsuperscript{\rm 1},
    Yi Sun\textsuperscript{\rm 1},
}
\affiliations{
    \textsuperscript{\rm 1}School of Information and Communication Engineering, Dalian University of Technology, China\\
    \textsuperscript{\rm 2}Shenyang Institute of Automation, Chinese Academy of Sciences, China\\

    \{baiwenbin6,20201071515,c,318546410\}@mail.dlut.edu.cn,
    \{linxbo,lslwf\}@dlut.edu.cn,lijianwen@sia.cn
}

\usepackage{bibentry}

\begin{document}

\maketitle

\begin{abstract}
The inherent difficulty and limited scalability of collecting manipulation data using multi-fingered robot hand hardware platforms have resulted in severe data scarcity, impeding research on data-driven dexterous manipulation policy learning. To address this challenge, we present a hand-agnostic manipulation transfer system. It efficiently converts human hand manipulation sequences from demonstration videos into high-quality dexterous manipulation trajectories without requirements of massive training data. To tackle the multi-dimensional disparities between human hands and dexterous hands, as well as the challenges posed by high-degree-of-freedom coordinated control of dexterous hands, we design a progressive transfer framework: first, we establish primary control signals for dexterous hands based on kinematic matching; subsequently, we train residual policies with action space rescaling and thumb-guided initialization to dynamically optimize contact interactions under unified rewards; finally, we compute wrist control trajectories with the objective of preserving operational semantics. Using only human hand manipulation videos, our system automatically configures system parameters for different tasks, balancing kinematic matching and dynamic optimization across dexterous hands, object categories, and tasks. Extensive experimental results demonstrate that our framework can automatically generate smooth and semantically correct dexterous hand manipulation that faithfully reproduces human intentions, achieving high efficiency and strong generalizability with an average transfer success rate of 73\%, providing an easily implementable and scalable method for collecting robot dexterous manipulation data. 
\end{abstract}



\section{Introduction}

Transferring human capabilities in flexible object manipulation to multi-fingered dexterous robotic hands has long been a core topic in robotics. Dexterous manipulation transfer is crucial for two reasons: (a) transferring human skills into robotic productivity, allows robots to use everyday tools to better assist human life; (b) generating manipulation data, facilitates the development of data-driven robotic systems. Compared to real-world transfer, simulation-based transfer offers advantages such as higher data efficiency, broader task scalability, and greater tolerance to failure. With the advent of high-fidelity physics simulators, models trained in simulation have demonstrated strong potential for real-world deployment \cite{wang2024cyberdemo,jiang2024dexmimicgen}, thereby fueling growing interest in simulator-based manipulation transfer \cite{chen2022dextransfer,qin2022dexmv,liu2023dexrepnet,wang2023physhoi,zhang2023learning,guzey2024bridging,chen2024object,liu2024parameterized,liu2025dextrack}. However, replicating human-level manual dexterity remains a significant challenge.

\begin{figure}[t]
    \centering
    \includegraphics[width=1\linewidth]{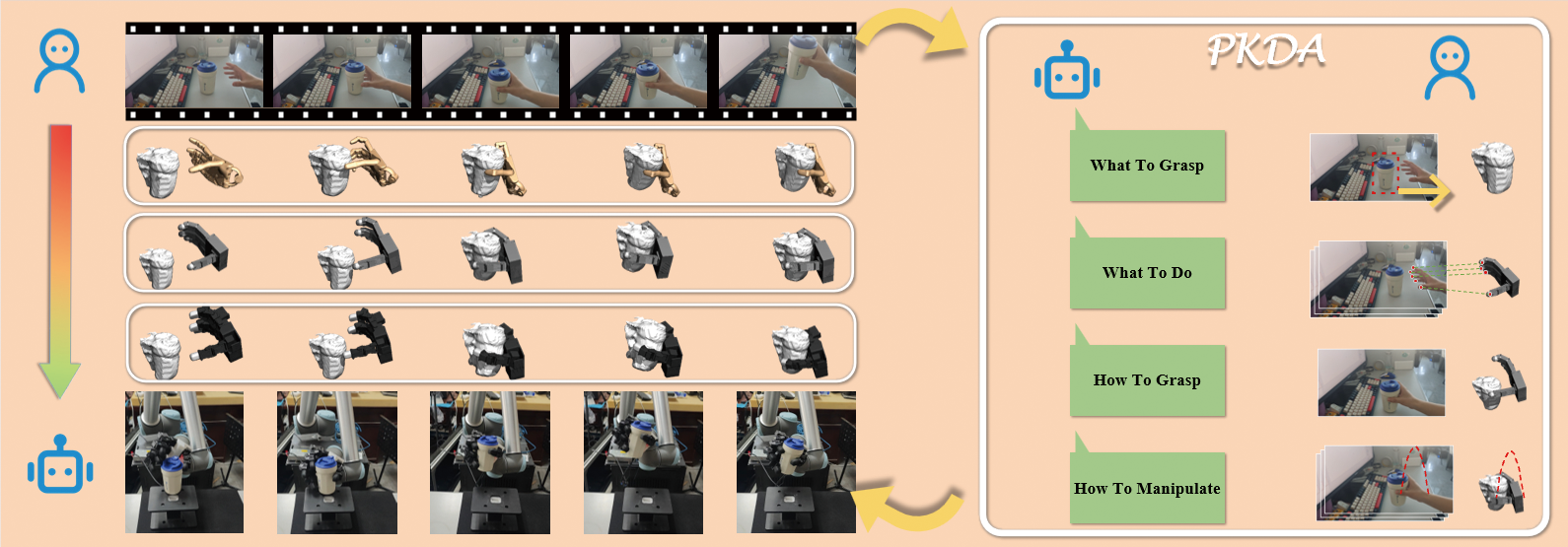}
    \caption{PKDA Overview: Starting from human video demonstrations, human manipulation is mapped to various dexterous hands. The system achieves perception-control integration for dexterous manipulation transfer by addressing four core questions: “what to grasp, what to do, how to grasp, and how to manipulate.”}
    \label{fig:enter-label1}
\end{figure}

Three key challenges hinder dexterous manipulation transfer: 1) Structural differences between human and robotic hands pose significant challenges for accurate motion retargeting. 2) Complex hand-object contact dynamics impede behavior transfer \cite{liu2024parameterized,pang2023global}.
3) The diversity of manipulation tasks limits the generalization of optimization frameworks. \cite{arunachalam2023holo,arunachalam2023dexterous,Park2025LearningTT,wang2024dexcap,qin2022one,handa2020dexpilot,qin2023anyteleop,sivakumar2022robotic} utilize cameras or wearable devices to facilitate the real-time transition from human manipulation to dexterous hand movements with human visual feedback, but is costly and challenging to scale up. Offline transfer techniques, as referenced in \cite{qin2022dexmv,chen2022dextransfer,liu2023dexrepnet,Li2024OKAMITH}, only employing kinematics mapping between hands without considering hand-object contact dynamic optimization, fail to adhere to physical constraints. While reinforcement learning (RL) methods \cite{rajeswaran2017learning,Christen2021DGraspPP,mandikal2022dexvip,dasari2023learning,guzey2024bridging,chen2024object,zhang2023learning} can explore interactions through autonomous trial and error, their inefficient exploration and task-specific reward design limit their generalizing ability.

To this end, we propose Dexterous Manipulation Transfer system via
 {\bf{P}}rogressive  {\bf{K}}inematic- {\bf{D}}ynamic  {\bf{A}}lignment (PKDA) framework, which is designed to be broadly adaptable to different dexterous hands, diverse manipulation tasks, and multiple object categories. As shown in Fig.1, the system only requires human manipulation RGB videos as input, and can automatically configure system parameters to transfer coherent control signals that drive various dexterous hands to complete diverse manipulation tasks.

Our main viewpoints are as follows: 1) The inherent coordination and interaction diversity existing in offline human demonstrations can yield valuable and reliable behavioral references. 2) The synergy of kinematics matching and RL can facilitate balancing the transfer efficiency and quality. Kinematic matching enhances RL by supplying high-quality initialization states and dependable exploration directions. Concurrently, RL investigates effective hand-object interactions under the guidance of kinematic matching and task-irrelevant rewards, forming a transfer path that is both kinematically aligned and dynamically optimized.

Based on the above perspectives, we model the dexterous manipulation transfer problem as a four-stage task: extracting hand-object priors, approaching the object, anthropomorphically grasping the object, and manipulating. Correspondingly, our PKDA framework contains four specially designed modules, as shown in Fig.2. \textbf{Interaction Perceptor} extracts hand-object interaction information. \textbf{Trajectory Proposer} maps human hand motion trajectories to the dexterous hands and generates primary control signals. \textbf{ContactAdapt Optimizer} employing RL, through thumb-guided pre-grasp initialization, action space rescaling, and unified reward, trains a residual policy, ensuring that the optimized trajectories meet both kinematic adaptability and physical interaction stability. \textbf{Wrist Trajectory Planner}, adjusts the wrist trajectory using the object trajectory as guidance, ensuring that the dexterous hand manipulation retains the task semantics. It is worth emphasizing that throughout the entire transfer process, it does not require task-specific parameter adjustment, enabling efficient adaptation to various manipulation tasks.

To validate the performance of our dexterous manipulation transfer system at different situations, we conducted extensive experiments in MuJoCo\cite{Todorov2012MuJoCoAP} across a wide variety of manipulation tasks, using three representative robotic hands: Adroit Hand, Allegro Hand, and Leap Hand. The proposed PKDA framework shows significant generalization ability, and achieves higher success rate, faster transfer speed than state-of-the-art methods.

Our contributions are summarized as follows:

\begin{itemize}
    \item We present a new system for transferring human hand manipulation to multi-fingered dexterous robot hand from human demonstration video. It exhibits excellent transfer stability across different dexterous hand configurations, manipulation tasks, and object categories through leveraging their commonalities. It provides an easily implementable, efficient and scalable technical solution.
    \item We propose the PKDA manipulation policy learning framework featured by a novel synergistic optimization of kinematics mapping and contact dynamics. Kinematics mapping guides imitation and restricts exploration of RL. Action space rescaling and thumb-guided pre-grasp initialization improve the efficiency of dynamic deviations correction. It shortens the transfer time while also ensures the successful transfer of manipulation capabilities.

\end{itemize}


\section{Related Work}
\subsection{Dexterous manipulation transfer}

Humans develop adaptive grasping skills through practice, and efficiently transferring these skills to dexterous hands can avoid inefficient learning from scratch, offering a promising path to enhance robotic manipulation \cite{liu2020skill,kadalagere2023review}. Many studies perform kinematic matching between human and robotic hands by establishing structural correspondence ( finger-wrist vectors \cite{qin2022dexmv,qin2023anyteleop}, interfinger vectors \cite{handa2020dexpilot}). However, due to the absence of contact dynamics optimization, such correspondences often require manual intervention to achieve successful physical interactions. Deep reinforcement learning (DRL) has shown strong capabilities in dynamic tasks like in-hand manipulation \cite{rajeswaran2017learning, OpenAI2019SolvingRC, Chen2021ASF, Chen2022VisualDI}, inspiring DRL-based transfer methods \cite{Christen2021DGraspPP, dasari2023learning, wang2023physhoi}, while effective, they suffer from task-dependent complex rewards design and inefficient exploration in high-dimensional space. Recent work seeks to unify kinematic and dynamic alignment \cite{chen2022dextransfer, liu2023dexrepnet, guzey2024bridging, liu2025dextrack, liu2024parameterized, yin2025dexteritygen, chen2024object, zhao2024dexh2r, lum2025crossing, park2025learning, li2025maniptrans}. Approaches like \cite{liu2023dexrepnet, chen2022dextransfer} combine retargeting with correlated sampling and local trajectory optimization, but random exploration still risks physically implausible grasps. \cite{li2025maniptrans} employs a two-stage RL pipeline but heavily depends on large, precisely labeled datasets and is vulnerable to noise. Differently, our method provides a new transfer scheme with the constrained RL exploration through multiple designs of action space rescaling, unified reward and thumb-guided pre-grasp initialization, improving efficiency, accuracy, and robust generalization.

\subsection{Demonstration data collection}

Demonstration data has been shown to significantly accelerate manipulation policy learning and improve generalization. Precisely recorded human hand manipulation using multiple cameras or motion capture systems are important data sources \cite{chao2021dexycb, kwon2021h2o, fan2023arctic, li2024favor}, but anatomical and kinematic differences hinder direct transfer to robotic hands.
A direct alternative is to collect data using the dexterous robot itself in physical environments \cite{cheng2024open, Park2025LearningTT, fang2023rh20t, zitkovich2023rt, khazatsky2024droid}, but high hardware costs and complex setups limit scalability. Thanks to advances in high-fidelity simulation and Sim2Real transfer, simulation data has proven effective for real-world tasks \cite{wang2024cyberdemo, jiang2024dexmimicgen, handa2023dextreme}. This has motivated the use of simulators for data collection, typical representatives including teleoperation \cite{qin2022one,park2024dexhub} and imitation learning \cite{wang2024dexcap}. Our method provides an efficient way to generate dexterous manipulation data that aligns with human manipulation in both kinematic and dynamic terms without the need for online human involvement.

\begin{figure*}[t]
    \centering
    \includegraphics[width=1\linewidth]{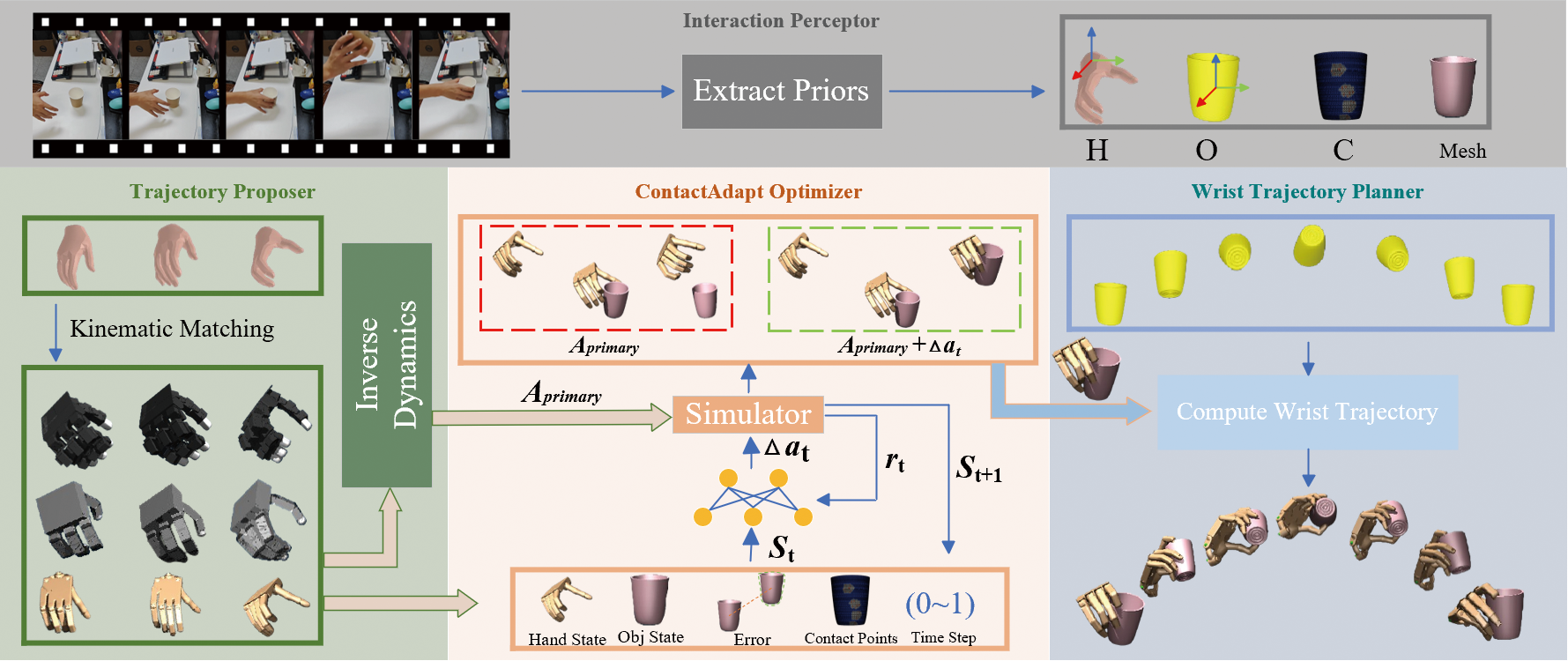}
    \caption{
    The PKDA system comprises four modules. Interaction Perceptor extracts key manipulation cues, including human hand posture H, object pose O, contact point C, and object mesh, from human demonstration videos. Trajectory Proposer retargets human hand movements into dexterous hand joint angle sequences, generating a primary control signal \(A_{primary}\) to guide the primary trajectory. However, lacking dynamic adjustment, this trajectory often leads to failed grasps (see red box). To improve grasp stability, the ContactAdapt Optimizer employs RL, where a residual policy modifies \(A_{primary}\) through \(\Delta a_{t}\). Wrist Trajectory Planner combines object motion with relative hand-object constraints to synthesize wrist trajectories and generate the complete control signal.}
    \label{fig:enter-label2}
\end{figure*}

\section{Method}
The PKDA pipeline of our system is illustrated in Fig.2, comprising four modules.

\subsection{A. Interaction Perceptor}
Interaction Perceptor undertakes the task of collecting the dynamic hand-object interaction information from the raw human manipulation videos, including hand trajectories \(H = \left \{h_{1},\dots,h_{t},\dots,h_{T}\right \} \), object trajectories \(O = \left \{  o_{1},\dots,o_{t},\dots,o_{T}\right \} \), with horizon length \(T\), and \(N\) contact points \(C = \left \{c_{1},c_{2},\dots,c_{N}\right \}\). \(h_{t} \in R^{18}\), represents the spatial positions of the fingertips (\(15D \)) and the palm orientation (\(3D \)). \(o_{t} =\left\{pos_t,ori_t\right\}\ \in R^{6}\), represents the pose of the object at the time t, where \(pos_t\) represents the 3D position of the object's center of mass, and \(ori_t\) represents the object's 3D orientation. \(c_{N} \in R^{N \times 3}\), represents the 3D coordinates of the contact points, and \(N\in \left \{2,3,4,5\right\}\) represents the number of fingertip contacts.

For datasets with known object models (e.g., DexYCB \cite{chao2021dexycb}, TACO \cite{liu2024taco}), we estimate hand and object pose trajectories using HFL-Net \cite{Lin_2023_CVPR},  designed for monocular RGB images with shared low/high-level features, separate mid-level backbones, and self-/cross-attention to enhance estimation accuracy. Contact points are firstly identified by computing the minimum fingertip-to-object distances and then retained those beneath a 5cm threshold.

For raw videos without ground truth object models, we adopt Hold \cite{fan2024hold} to jointly reconstruct 3D hand-object geometry. To mitigate impacts of object mesh defects caused by occlusion, we apply convex decomposition optimization to perform effective collision detection. To ensure physical plausibility, we also lower the reconstructed object's center of mass to improve stability.
 
\subsection{B. Trajectory Proposer}

The human hand has abilities of precise fingertip position control and adaptive postural adjustment. To mimic this, we employ a retargeting technique that maps a human manipulation trajectory \(H\) to a dexterous hand joint angle sequence \(Q=\left \{ q_{1},\dots,q_{t},\dots q_{T} \right \} \), using fingertip positions and palm orientation as constraints. \(q_{t} \in R^{D}\)  represents joint angles at time \(t\), and its dimension \(D\) depends on hand type (Appendix A.1 for details).

Existing methods align fingertip-to-wrist vectors between human and robotic hands \cite{qin2023anyteleop, qin2022dexmv}, emphasizing global pose imitation. However, due to hand size discrepancies, this approach introduces noticeable fingertip errors—especially problematic in autonomous manipulation without human feedback.

In our scheme, there is no human visual feedback at all. A decline in fingertip positioning precision alters contact points and contact directions on object surface, risking compromised grasp stability and consequently reducing grasp success rate. To address this, we use the fingertip position vectors in the world coordinate system as the main objective and the palm orientation as an auxiliary constraint for kinematics mapping. This decouples fingertip localization from wrist alignment, reducing sensitivity to morphological differences. Inspired by \cite{handa2020dexpilot}, we formulate the mapping as a nonlinear optimization problem that minimizes the position errors between human and robotic fingertips.

The objective function is defined as follows, including fingertip position constraint \(E_f\), palm orientation constraint \(E_o\), and temporal smoothing constraint\(E_s\).
\begingroup
\small
\begin{equation}
\min_{\mathbf{q}_{t}} \left(w_{f} E_{f}+w_{o} E_{o }+w_{s} E_{s}\right)
\end{equation}
\begin{equation}
E_{f}=\sum_{i=1}^{K}\left\|\mathbf{v_{i}^{H}}\left(\mathbf{H_{t}}\right)- \mathbf{v_{i}^{R}}\left(\mathbf{q}_{t}\right)\right\|^{2},        
\end{equation}
\begin{equation}
E_{o}=\mathbf{G}\left(\mathbf{M}_{t}^{H}, \mathbf{M}_{t}^{R}\right),   
\end{equation}
\begin{equation}
E_{s} = \left\|\mathbf{q_{t}}-\mathbf{q_{t-1}}\right\|^{2},              
\end{equation}
\endgroup

\(\mathbf{v}_{\mathbf{i}}^{\mathbf{H}}\) and \(\mathbf{v}_{\mathbf{i}}^{\mathbf{R}}\) in Eq.2 represent the fingertip position vector of the \(i\) th finger of the human hand and the corresponding finger of the dexterous hand, \(\mathbf{v}_{\mathbf{i}}^{\mathbf{R}}\) is computed by the forward kinematic model of the dexterous hand. { \bf{G}} in Eq.3 is the minimum geodesic distance\cite{huynh2009metrics} which is used for calculating the difference between the palm  orientation of the human hand \(\mathbf{M}_{t}^{H}\) and that of the dexterous hand \(\mathbf{M}_{t}^{R}\). Eq.4 is a temporal smoothing term to penalize the occurrence of large joint angle changes in adjacent action frames. The obtained discrete joint angle sequence \(Q\) is converted into the  control sequence \(A_{primary}\) using  an inverse dynamics-based joint angle-control signal conversion method \cite{qin2022dexmv}. For a dexterous hand with K fingers, the corresponding relation of fingers, as well as the definition of the palm orientation are presented in Appendix A.1.

\subsection{C. ContactAdapt Optimizer}

Simple motion mapping can imitate grasp postures but fail to convey force closure and dynamic contact strategies, reducing grasp stability under disturbances. To address this, we use RL to optimize grasp dynamics, enabling human-like contact adaptation. However, task diversity complicates reward design, and large action spaces hinder policy efficiency. We propose: 1) RL-Configurator, a cross-task feature extractor for unified reward design. 2) action space rescaling to mitigate ineffective exploration. 

\noindent{\bf RL-Configurator:} In cross-task settings, task specificity poses a significant challenge to the unified RL training paradigm. The core objective of RL-Configurator is extracting common features from object poses, primary trajectories of the dexterous hand, and fingertip contact points across different tasks, thereby establishing a unified RL configuration. As shown in Fig.3 (left), it regulates the configuration of the RL task from the following two main perspectives: 

1. Initialize the pre-grasp state for RL based on \(A_{primary}\). Considering the thumb's dominant role in the whole grasping process, making it approach the optimal contact position at the early stage of the interaction will significantly reduce the difficulty of policy exploration. Therefore, based on the real-time collision detection and the position detection of thumb tip, we choose the state where the hand-object are not in contact and the thumb tip is closest to the corresponding grasp point as the initial state, and we call this state as the pre-grasp state\(( \hat{q}_{pre},\dot{q}_{pre})\), where \(\hat{q}_{pre}\) and \(\dot{q}_{pre}\) represent the joint angle and joint velocity.

2. Set the goal for RL. Since most manipulations rely on grasping objects as a prerequisite, we abstract the initial phase of all manipulation tasks as pick-up. The objective is to bring the object to the target pose \(o_{target}\) (the pose where the object deviates from its initial position by 0.1m for the first time in its trajectory).

\noindent{\bf Action space rescaling:} The action space is delineated as \(a \in{R}^{D}\), where the preceding six dimensions control wrist translation and rotation, while the remaining dimensions control the finger posture of the dexterous hand. To address the distinct learning characteristics and task requirements of wrist and finger movements, we employ action space rescaling during the RL process, as shown in Fig.3 (right). We compress the wrist joint movement range from the global workspace to a local neighborhood \(\mathcal{N}\left(\hat{q}_{{pre}}, \rho \right)\) centered around the wrist component of the pre-grasp joint angle \(\hat{q}_{pre}\), while maintaining the full range of motion for the finger joints. This constraint not only encourages diverse interaction attempts between the fingers and objects, reducing ineffective exploration of the policy, but also prevents over-shooting behaviors caused by excessive wrist motor signal.

\begin{figure}[t]
    \centering
    \includegraphics[width=1\linewidth]{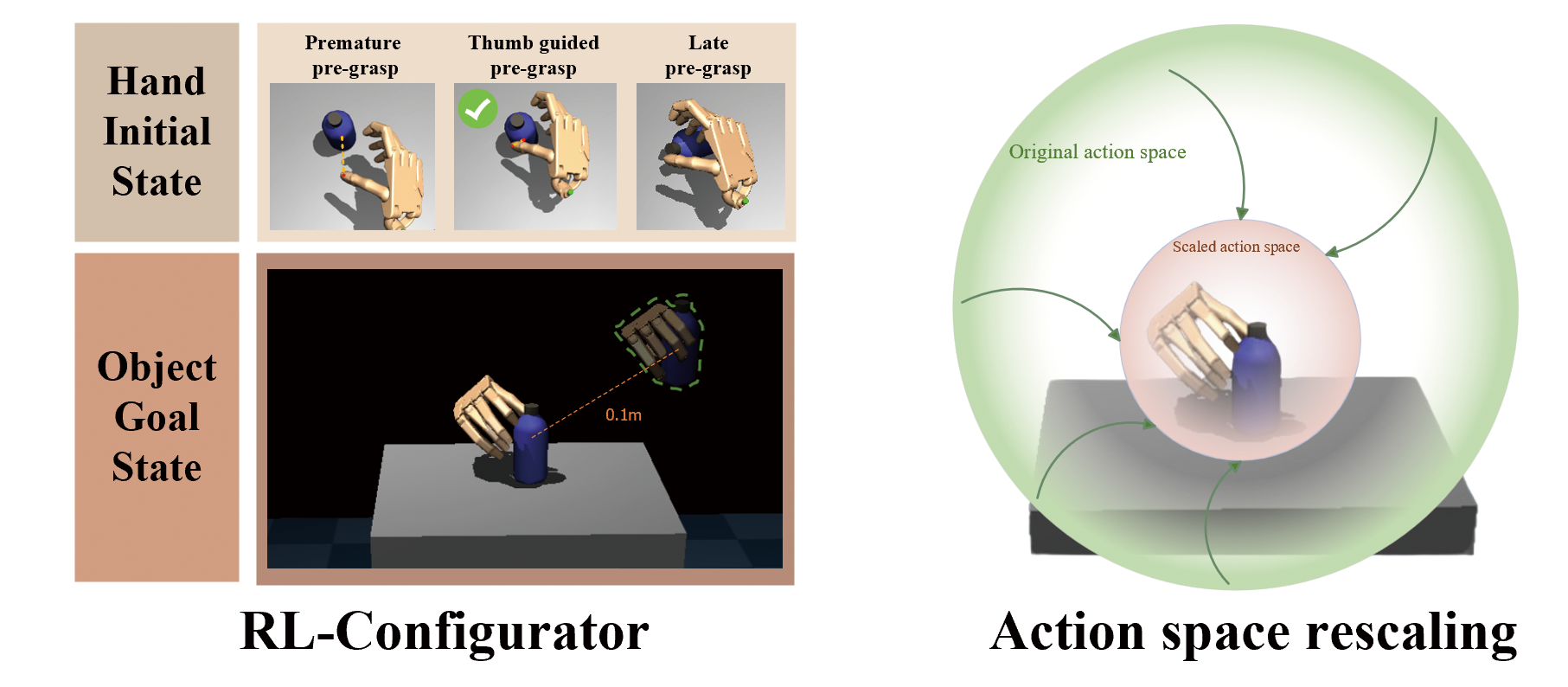}
    \caption{RL-Configurator (left) standardizes diverse tasks by configuring pre-grasp and object goal state. The Action Space Rescaling module (right) compresses wrist motion space into the neighborhood of the pre-grasp to enable efficient hand-object interaction exploration. }
    \label{fig:enter-label3}
\end{figure}

\noindent{\bf Reward Design:} To effectively train grasping policies across diverse object types, we implement a hierarchical and unified reward with three key components: \textit{1) Approach reward} ($r_{approach}$): It guides fingertips toward target contact points, providing positive feedback when fingertips approach their designated positions. \textit{2) Grasp reward }($r_{grasp}$): Activating when all fingertips are within the contact tolerance threshold ($\varepsilon = 0.06m$). It incorporates both contact reward \(r_{con}\) and imitation reward \(r_{sim}\). The contact reward encourages multi-point contact between the fingertips and the object through collision detection. The imitation reward promotes hand posture imitation by computing the cosine similarity between the current joint angles and the target joint angles obtained through retargeting. \textit{3) Lifting reward} ($r_{lift}$): Activating when the thumb and one other finger make contact with the object. It guides the dexterous hand to maneuver the object to the target pose. The complete reward function combines these components (Appendix A.3 for more details). The RL training of the grasping policy is presented in Appendix A.4.

\subsection{D. Wrist Trajectory Planner}

For dexterous manipulation transfer tasks, maintaining the consistency of action intentions before and after transfer is an important requirement. Taking the task of drinking water with a cup in hand as an example, the most important requirement of the manipulation transfer is that the dexterous hand can hold the cup to realize the dynamic process of ``rise-tilt (drink) -put down''. In order to meet such basic requirements, we designed a wrist trajectory planner guided by the dynamic pose change of the object. The interaction between the dexterous hand and the object is modeled as a low dynamic manipulation, that is, there is no relative sliding after the hand picks up the object. We extract the trajectory of the object in the manipulation stage \(\left \{o_{grasp} ,\dots,o_{t},\dots ,o_{T}\right \}\), and the pose of the wrist of the dexterous hand in the stable grasping \(T_{grasp}\), where \(o_{grasp}\) refers to the object pose corresponding to the end time step of RL. According to the computed wrist trajectory \(T_{t }=o_{{t}} \cdot\left(T_{{grasp }}^{-1} \cdot o_{{grasp }}\right)^{-1}\) at each time step, the PD controller is used to control the motion of the dexterous wrist.

\section{Experiments}

\subsection{A. Experiment Settings}

To fully test our approach, we constructed test datasets from the following multiple scenarios.

\begin{itemize}
\item  \textit{Full-Information Scenario:}
It enables access to accurate hand poses, object poses, and hand-object contact annotations. We manually selected about 600 trajectories involving single-hand interactions from GRAB dataset \cite{Taheri2020GRABAD} for evaluation.

\item  \textit{Model-Known Visual Scenario:}
Hand and object poses, as well as contact points, are estimated from monocular RGB videos, assuming the object’s 3D model is known. 10 right-hand manipulations involving geometrically diverse objects were selected from each dataset ( DexYCB \cite{chao2021dexycb} and TACO \cite{liu2024taco}) for evaluation. 

\item  \textit{Model-Unknown Visual Scenario:}
Only RGB videos are available. This scenario reflects more actual situations. Videos are shot with our own color camera,  featuring 5 diverse manipulation tasks using common household objects.
\end{itemize}

To quantitatively evaluate the performance of our method against the baseline, we adopt four key metrics: 
\textit{1) SR Grasp(\%)}: Grasping success rate. Successful grasping means the object can be held within 0.05m of the target position. \textit{2) SR Follow(\%)}: Following success rate. Successful following means dexterous hand can hold the object stably without dropping it throughout the entire manipulation procedure. \textit{3) Er(°)}: Deviation of object rotation, computed by geodesic distance \(G\), \(E{r}=\frac{1}{T} \sum_{i=1}^{T}\mathbf{G}( ori_{i}, \hat{ori_{i}})\). \textit{4) Ep(m)}: Deviation of object translation, computed by Euclidean distance \(d(x,y) =  \left\| \mathbf{x} - \mathbf{y} \right\|_{2}\), \(E{p}=\frac{1}{T} \sum_{i=1}^{T}d(pos_{i},\hat{pos_{i}})\).

Furthermore, we introduce a new metric, \textit{Transfer Success Rate (TSR(\%))}, to assess intention-level consistency (Appendix B.1 for details).

\subsection{B. Comparative Experiments}

We use the following methods as the baselines for comparison:

\textit{1. Anyteleop} \cite{qin2023anyteleop}: Anyteleop achieves the alignment of the human hand and the dexterous hand through the hand pose retargeting.

\textit{2. PGDM} \cite{dasari2023learning}: PGDM uses pre-grasp poses from multiple sources (motion capture, expert teleoperation,manual annotations) to initialize the RL environment and designs rewards based on reference trajectory reproduction for precise tracking. In the comparative experiment, the manipulation trajectory is directly generated utilizing the pre-trained policy provided by the author.

\textit{3. D-Grasp} \cite{Christen2021DGraspPP}: D-Grasp generates physically plausible dynamic interactions using single-frame grasping pose as reference and specially designed rewards. we re-implemented D-Grasp's reward design in MuJoCo (originally RaiSim)  to eliminate the potential deviation caused by different simulator. Implementation details are in Appendix B.2.

Since the implementation of the PGDM method relies on the pre-extracted pose provided by the authors, which is only available on the TCDM benchmark they provided, we only evaluated the subset of tasks where the TCDM benchmark overlaps with our selected data from the GRAB dataset, totaling 40 manipulation tasks. The task list is provided in Appendix B.3.

\begin{table}[t]
\centering
\small
\begin{tabular}{@{}lllll@{}l} \hline
                                  & \multicolumn{1}{c}{SR Grasp ↑} & \multicolumn{1}{c}{SR Follow ↑} & \multicolumn{1}{c}{Ep ↓} & \multicolumn{1}{c}{Er ↓} & TSR↑\\ \hline
\textit{Anyteleop}&                                 12.5\%&                                  7.5\%&                            N/A&                             N/A&7.5\%\\
\textit{PGDM}          &                                 72.5\%&                                  72.5\%&                            0.005&                             18.2&72.5\%\\
\textit{D-Grasp}       &                                 62.5\%&                                  60\%&                            0.067&                             35.5&57.5\%\\
\textit{PKDA-P}&                                 80\%&                                  80\%&                            0.058&                             31.5&77.5\%\\
\textit{PKDA-F}&                                 84.2\%&                                  77.6\%&                            0.060&                           
 34.8&73.3\%\\ \hline
\end{tabular}
\caption{Results of different methods under the full information scenario using the Adroit Hand. PKDA-P represents the results tested on TCDM Task(40 sequences), and PKDA-F represents the results on GRAB (600 sequences).}
\label{tab:my_table}
\end{table}
\begin{figure}[t]
    \centering
    \includegraphics[width=1\linewidth]{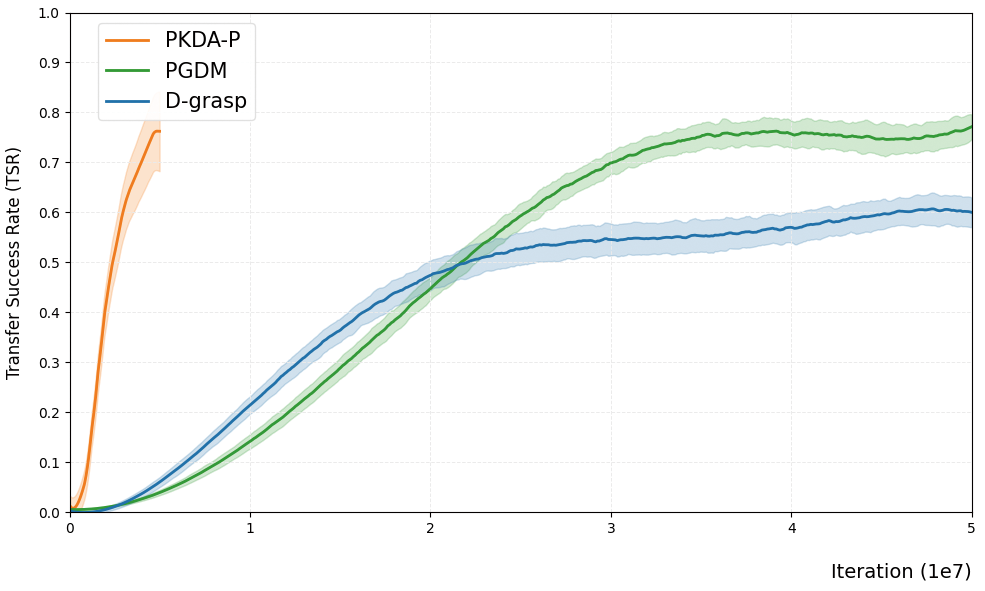}
    \caption{Learning efficiency comparison on the TCDM task using the Adroit Hand. }
    \label{fig:enter-label4}
\end{figure}
\begin{figure}[t]
    \centering
    \includegraphics[width=1\linewidth]{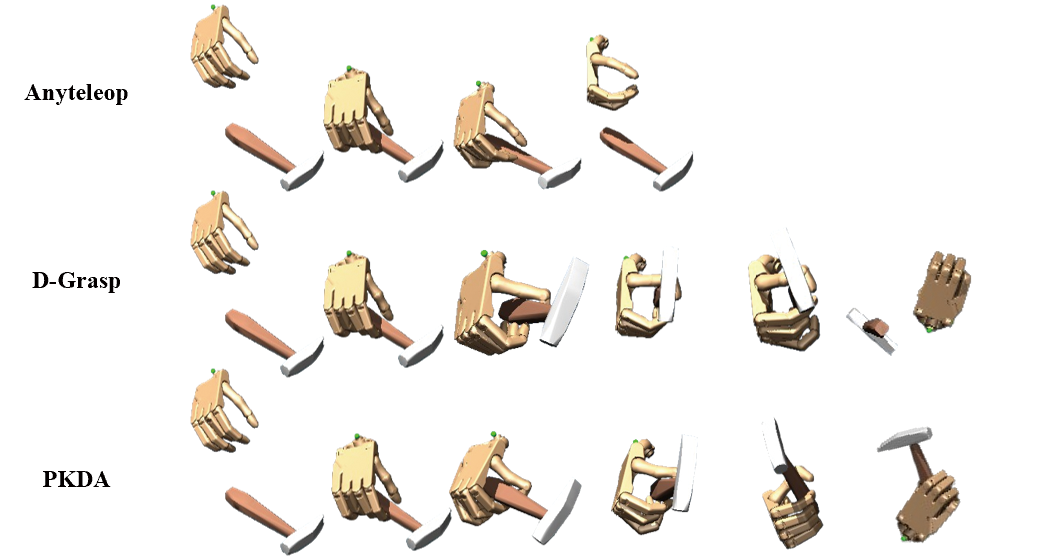}
    \caption{Comparison of the transfer quality on repetitive task of knocking a nail.}
    \label{fig:enter-label5}
\end{figure}

As shown in Table 1, our method outperforms baselines on SR Grasp, SR Follow and TSR. Tracking accuracy is better than D-Grasp but lower than that of PGDM method. PGDM takes the object trajectory as a strong constraint, sacrificing transfer efficiency to exactly reproduce trajectories. We argue that prioritizing transfer of manipulation action intentions should be more reasonable in terms of practical goal and challenges. Hence we optimize only the grasp phase with RL and use a PD controller for the remaining motion. This yields markedly higher transfer efficiency (Fig.4). In addition, PGDM only transfers a partial trajectory starting from the pre-grasp, whereas our method transfers a complete manipulation sequence, including approaching the object and putting the object back at the end of task. The Retarget-only method Anyteleop struggles to resist inertial disturbances, often resulting in ``lifting and slipping down'', it occasionally succeeds with objects like goblets due to favorable lifting postures, highlighting the need for dynamic optimization. D-Grasp, guided by static poses, excels at relocation but falters in repetitive tasks like knocking a nail (Fig.5). Overall, our approach offers both higher learning efficiency and superior transfer success.

\subsection{C. Robustness Evaluation}

\begin{figure}[t]
    \centering
    \includegraphics[width=1\linewidth]{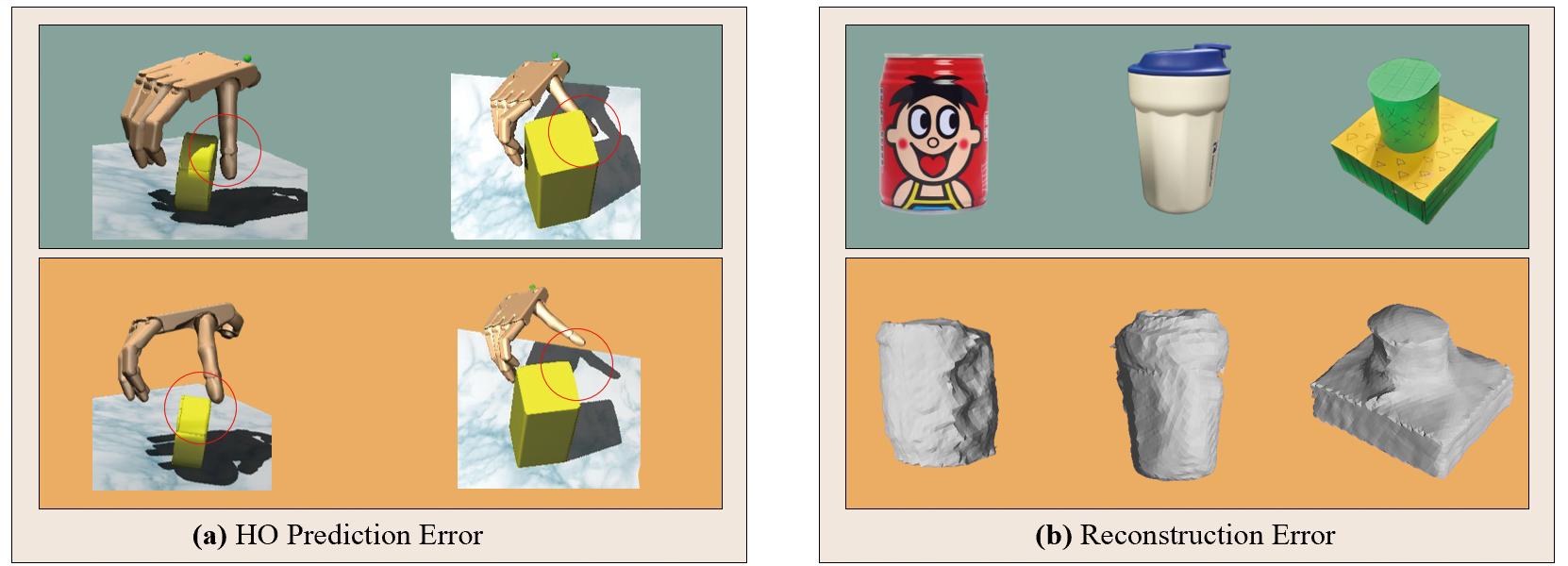}
    \caption{Perception inaccuracies: (a) hand pose (upper: ground truth, lower: estimation result), (b) reconstruction defects (upper: 2D image, lower: 3D mesh). }
    \label{fig:enter-label6}
\end{figure}

To assess robustness under non-ideal conditions with visual perception errors, we conducted experiments in both Model-Known and Model-Unknown Visual Scenarios. Two common visual perception inaccuracies occur, including pose estimate errors and object reconstruction defects. Fig.6 illustrates these shortcomings: (a) shows that pose estimation error yields unsatisfactory pre-grasp pose; (b) displays object reconstruction defects, where the reconstructed object model retains only a crude shape showing large error in details.

\begin{table}[b]
\centering
\small
\setlength{\tabcolsep}{3pt} 
\renewcommand{\arraystretch}{1.1} 
\begin{tabular}{lccccc}
\hline
\textbf{} & SR Grasp↑ & SR Follow↑ & Ep↓ & Er↓ & TSR↑ \\
\hline
\textbf{Model-Known} & 75\% & 70\% & 0.034 & 22.4 & 70\% \\
\textbf{Model-Unknown}  & 80\% & 80\% & 0.033 & 34.8 & 80\% \\
\hline
\end{tabular}
\caption{Experimental results of PKDA in the scene with perception inaccuracies using the Adroit Hand.}
\label{tab:vision_adroit}
\end{table}

As shown in Table 2, the success rates are no less than 70\% in the two scenarios. The results verify that the PKDA framework enables resistance to interference, so that a robust policy can still be learned under the condition of input uncertainty. The visualization results are presented in Fig.9 in Appendix B.4.

\begin{table}[t]
\centering
\small
\setlength{\tabcolsep}{3pt} 
\renewcommand{\arraystretch}{1.1} 
\begin{tabular}{lccccc}
\hline
\textbf{} & SR Grasp↑ & SR Follow↑ & Ep↓ & Er↓ & TSR↑ \\
\hline
Adroit & 80\% & 80\% & 0.0584 & 31.5 & 77.5\% \\
Allegro & 77.5\% & 72.5\% & 0.0569 & 32.8 & 72.5\% \\
Leap    & 70\% & 67.5\% & 0.0544 & 31.7 & 67.5\% \\
\hline
\end{tabular}
\caption{Result of PKDA-P on different dexterous hands.}
\label{tab:hand_compare}
\end{table}
\begin{table}[t]
\centering
\setlength{\tabcolsep}{4pt} 
\renewcommand{\arraystretch}{1.1} 
\begin{tabular}{lcc}
\hline
\textbf{Method} & \textit{Finger-wrist Vector} & \textbf{Ours} \\
\hline
TSR↑ & 70\% & 77.5\% \\
\hline
\end{tabular}
\caption{TSR of PKDA-P with different retargeting methods using the Adroit Hand.}
\label{tab:redirect_compare}
\end{table}
\begin{table}[b]
\centering
\small
\setlength{\tabcolsep}{5pt} 
\renewcommand{\arraystretch}{1.1} 
\begin{tabular}{ccccccc}
\hline
        & TG+R& IG+R& MG+R& TG& IG& MG\\
\hline
Nearest & 77.5\%& 70\%& 67.5\%& 37.5\%& 32.5\%& 25\%\\
0.05m   & 72.5\%& 70\%& 65\%& 42.5\%& 42.5\%& 30\%\\
0.1m    & 67.5\%& 65\%& 67.5\%& 45\%& 40\%& 30\%\\
\hline
\end{tabular}

\caption{TSR of PKDA-P under different pre-grasp selection strategies, with and without action space rescaling. TG, IG, and MG represent Thumb-Guided, Index-Guided, and Middle-Guided strategies, respectively. ``+R'' denotes the use of action space rescaling. All experiments are conducted using the Adroit Hand. }
\label{tab:yourlabel}
\end{table}

\subsection{D. Experiments on Different Dexterous Hand}

To validate PKDA's cross-hand capability, we test three representative dexterous hands, with results in Table 3. Our method achieves effective transfer across diverse hand configurations. Adroit Hand performs best with a 77.5\% transfer success rate. Larger hands like Allegro and Leap face challenges on small or slender objects (e.g., hammers), lowering their success rates. Despite significant differences in joint freedom, finger length, and kinematics, position errors (0.054–0.058) and rotation errors (31°–33°) remained consistent, demonstrating steady transfer quality. This consistency stems from preserving hand structural decoupling: fingertip mapping relies on spatial geometry rather than topology, and wrist trajectories are driven by relative hand-object relations. Our method prioritizes functional alignment over hardware specifics, enabling seamless adaptation to different hands by only mapping finger correspondences without adjusting overall parameters. The thumb-guided pre-grasp posture design ensures consistent contact strategies across hands, facilitating subsequent learning. In Appendix B.4, Fig.10 displays pre-grasp postures autonomously generated for Leap, Allegro, and Adroit, and Fig.11 displays manipulation sequence. 

\subsection{E. Ablation Experiments}

We construct the following ablation experiments to quantify the contribution of each core component.

\textit{1. Retargeting method:}
Aim to investigate the influence of retargeting method on the PKDA. We specifically investigated two retargeting methods: one guided by the finger-wrist vector and the other by the absolute position of the fingertip. The remaining components of the transfer procedure aligned with the PKDA.

\textit{2. Pre-grasp pose selection strategy:}
Explore two core aspects for obtaining high-quality pre-grasp poses:

(a) Guidance Condition: PKDA uses the thumb and its corresponding contact point. For comparison, we test index finger and middle finger guidance.

(b) Triggering moment: Under the premise of avoiding collisions between the hand and the object, PKDA selects the triggering moment that reaches the closest finger-to-object distance (Nearest). We also evaluate triggering at preset distance thresholds (0.05/0.1 m).

\textit{3. Action space rescaling:}
Performing diversified manipulation tasks with a unified model requires larger workspace, indicating that the dexterous hand has to spend more exploration time to seek for the correct action. To improve exploration efficiency, we propose action space rescaling mechanism in PKDA framework. The ablation experiment will validate its necessity.

Ablation results for the retargeting method (Table 4) show that using fingertip-wrist vectors reduces TSR by 7.5\%. Table 5 reveals that index/middle finger guidance reduces TSR by 7.5\%/10\% versus thumb guidance under Nearest conditions. When using thumb guidance, increasing preset distance thresholds lowers TSR (e.g., dropping to 67.5\% at 0.1m vs. Nearest). Crucially, action space rescaling has the most significant influence: Taking the ``thumb-guided'', ``Nearest'' triggering moment as an example, removing this mechanism resulted in obvious performance degradation, with the TSR falling to 37.5\%. Observations during experiments reveal that without action space rescaling, the dexterous hand exhibites unintended overshooting behavior during grasping. These results validate that rescaling is essential for suppressing detrimental action space interference during policy learning. 

\subsection{F. Real-World Experiments}

To validate the feasibility of our generated manipulation trajectories, we conduct real-world experiments on a UR10 robot arm equipped with a Leap Hand (Fig.7), performing three representative tasks—Shake, Pour, and Stamp—using daily objects. The trajectories generated in simulation are directly executed in the real world via open-loop control, successfully reproducing the intended manipulations. Implementation details are provided in Appendix C. 

\begin{figure}[t]
    \centering
    \includegraphics[width=1\linewidth]{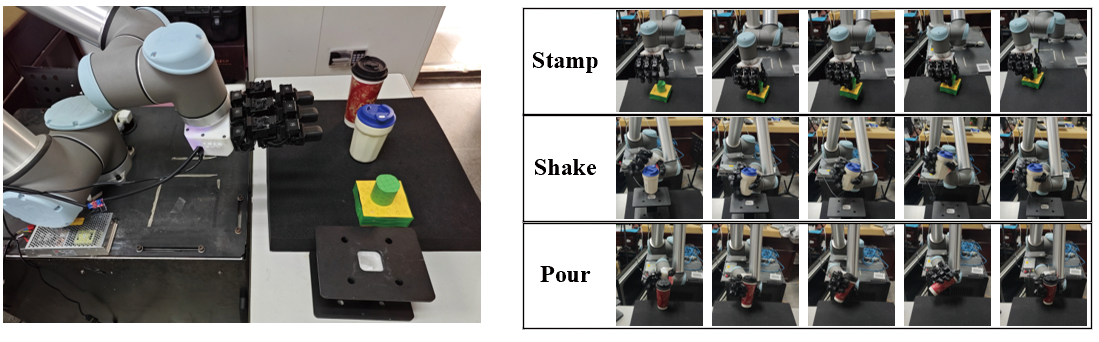}
    \caption{Real-world experiment. The left image shows the experimental scene configuration, while the right image displays the dexterous manipulation sequence deployed on the hardware.}
    \label{fig:enter-label7}
\end{figure}

\section{Summary And Limitations}

We propose PKDA, a framework for transferring dexterous manipulation from human demonstration videos. It extracts reference trajectories using pose estimation and object reconstruction, then maps them to dexterous hand motions through kinematic matching. A residual RL policy, guided by a unified reward, further optimizes hand–object contact dynamics. Wrist trajectories are then computed from object motion to preserve the full manipulation intent. Experiments across various dexterous hands and tasks show that PKDA outperforms baselines in both success rate and learning efficiency.

Limitations:
Although PKDA efficiently transfers manipulation sequences without large-scale pretraining, it primarily handles stable contact patterns. Future work will explore dexterous manipulation with dynamic multi-contact changes.

\section{Acknowledgments}
This work was supported by the National Nature Science Foundation of China.(62373075,61873046)

\bibliography{aaai2026}

\appendix

\section{Appendix A }

\subsection{A.1 The correspondence between hands}

The corresponding relationship of fingertips between the human hand and dexterous robot hand is shown in Fig.8 (left), where the same color indicates the corresponding fingertip. Different dexterous hands have different structures and joint degree of freedom, being 16 for Leap hand and Allegro hand, 24 for Adroit hand, respectively. The metacarpophalangeal joints of index finger and ring finger, and the wrist form the palm plane. The palm orientation vector is defined as the normal of the palm plane, depicted by the red arrow in Fig.8 (right). The palm orientation definition shows good stability in various mechanical hand configurations.

\begin{figure}[h]
    \centering
    \includegraphics[width=1\linewidth]{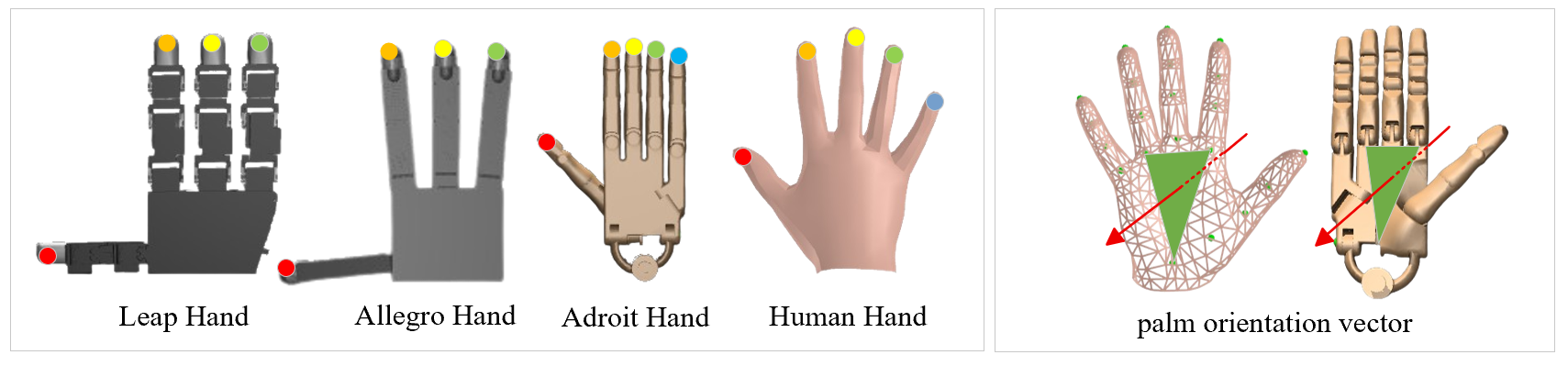}
    \caption{The correspondence setting of human hand and different dexterous hand. The left figure indicates the correspondence of fingertips, and the red arrow in the right figure indicates the palm orientation}
    \label{fig:enter-label8}
\end{figure}

\subsection{A.2 Joint angle-control signal conversion}

Many dynamic factors, such as joint inertia, actuator nonlinearity and environmental disturbance, will cause distorted actions performed by the dexterous hand according to the kinematic mapped trajectory. To amend the deviations, we use an inverse dynamics-based joint angle-control signal conversion method \cite{qin2022dexmv} to mitigate this adverse effect. First, to suppress sudden changes in motion, a third-order derivative minimization fitting is performed on the discrete joint angle sequence \(Q\) to generate a continuous trajectory function \(q(t)\). After that, the velocity \(q^{\prime}(t)\) and acceleration \(q^{\prime\prime}(t)\) are derived analytically from \(q(t)\) and incorporated into the inverse dynamics model \(\tau(t) = f_{inv}( q (t),{q}^{\prime}(t),q^{\prime\prime}(t))\) to compute the theoretical joint torque. Next, \(\tau(t)\) is converted into the requisite control signals \(a(t)\) by integrating the motor characteristics. \(a(t)\) is sampled into a control sequence \(A_{primary}\) at the sampling frequency \(F\). The joint offset arising from the self-weight of the dexterous hand is alleviated through gravity correction. Under the control of \(A_{primary}\), the motion trajectory produced by the dexterous hand effectively repeats the manipulation intentions, exhibiting an anthropomorphic hand pose to execute the actions of approaching, grasping, and manipulating an object. \(A_{primary}\) is named the primary control command, and the trajectory under its control is the primary trajectory \(PT = \left \{( \hat{q_{1}},\dot{q_{1}}),\dots ,( \hat{q_{t}},\dot{q_{t}}),\dots ( \hat{q_{T}},\dot{q_{T}})\right \} \).

\subsection{A.3 Details of rewards}

In ContactAdapt Optimizer, the unified reward for training grasping policies is hierarchical, including three terms that are approach reward (\(r_{approach}\)), grasp reward (\(r_{grasp}\)) and lifting reward (\(r_{lift}\)).

 {\bf{Approach reward}}

In order to guide the dexterous fingertip \(\mathbf{v_{i}^{R}}\) to quickly approach the corresponding contact point \(\mathbf{c_{i}}\), the reward function \(r_{{approach}}\) is designed as follows:
\begin{equation}
r_{ {approach}} =  \max(d_{  {closest}} - \sum_{ {i=1}}^{N} d(\mathbf{v_{i}^{R}},\mathbf{c_{i}}) , 0 )
\end{equation}
\(d_{  {closest}}\) records the historical minimum between the fingertips and the contact points during the current training cycle (initialized to \(-\infty \)). \(r_{appraoch}\) gives a positive reward when the fingertip approaches the contact point consistently. \(d()\) represents Euclidean distance, and \(N\) is the number of fingers.

 {\bf{Grasp reward}}

To prevent the occurrence of certain fingers contacting the object while others remain distant from the contact point on the object's surface and to encourage the dexterous hand to make a coordinated and synchronous contact with the object, the grasp reward is activated only when all fingertips enter the contact tolerance range \(\varepsilon\) (set to 0.06m). The  activation condition of the grasp reward \(r_{  grasp}\) is CE ( {\bf{C}}lose  {\bf{E}}nough):

\begin{equation}
CE = 
\left\{
\begin{array}{ll}
1, & \mbox{if } \forall i \in \{1,\dots,N\},\ d(\mathbf{v}_{i}^{R}, \mathbf{c}_{i}) \le \varepsilon \\
0, & \mbox{otherwise}
\end{array} 
\right.
\end{equation}

The grasp reward is defined as follows:
\begin{equation}
r_{\mathrm{con}} = \sum \limits_i \mathbf{1}(\mathbf{f_d}, \mathbf{o_s})
\end{equation}
\begin{equation}
r_{sim} = \frac{\mathbf{q}_t \cdot \mathbf{q}_{ {target}}}{\left\| \mathbf{q}_t \right\|_{2} \left\| \mathbf{q}_{ {target}} \right\|_{2}}
\end{equation}
\begin{equation}
r_{ {grasp}} = \beta_{ {con}}\times  
 r_{con}+\beta_{ {sim}}\times r_{sim}
\end{equation}
\begin{equation}
      \mathbf{1}(i, j)=\left\{\begin{array}{ll}
        1, &   {if} \  d(i,j) \leq \phi \\
        0, &   { otherwise }
        \end{array}\right.
 \end{equation}

The grasp reward \(r_{ {grasp}}\) comprises two components: the contact reward \(r_{con}\) and the imitation reward \(r_{sim}\). The contact reward promotes multi-contact interaction to improve force closure stability, quantified by the number of collisions between the distal phalanx \(f_d\) of the dexterous hand and the object's surface \(o_s\). \(\mathbf{1}\) is a Boolean variable, determined by assessing whether the shortest distance between the distal phalanx and the object is below the threshold \(\phi = 0.002m\). The imitation reward utilizes the joint angle acquired by retargeting as the target value \(q_{ {target}}\). It encourages the formation of human-like grasping configurations, as well as simultaneously regulating finger pose that has no corresponding contact points. For instance, a five-fingered hand may utilize only two fingers to manipulate an object, and the incorporation of imitation reward can guide the remaining three fingers to emulate human hand posture. The grasping reward \(r_{con}\) and the imitation reward \(r_{sim}\) are weighted by the parameters \(\beta_{ {con}}\) and \(\beta_{ {sim}}\).

 {\bf{Lifting reward}}

In order to achieve stable grasping behavior, the thumb, which serves as the primary contact finger, must be in contact with the object. Additionally, a secondary contact point must be established by at least one auxiliary finger to balance the grasping torque. In accordance with \cite{Qin2022DexPointGP}, we establish the foregoing conditions to activate the lifting reward and represent it using HT ( {\bf{H}}ave  {\bf{T}}ouched).

\begingroup
\small 
\begin{equation}
HT = 
\left\{
\begin{array}{ll}
1, & \mbox{if } \mathbf{1}( \mathbf{f}_t, \mathbf{o}_s ) \cdot \left( \sum_{i=1}^{N} \mathbf{1}(\mathbf{f}_i, \mathbf{o}_s )\right) \geq 1 \\
0, & \mbox{otherwise}
\end{array}
\right.
\end{equation}
\endgroup

\(\mathbf{1}{(f_t,o_s)}\) indicates whether the thumb is in contact with the object, and \(\mathbf{1}{(f_i,o_s)} \) indicates whether the other fingers except the thumb are in contact with the object.

The lift reward \(r_{lift}\) encourages the dexterous hand to maneuver the object to the target pose:

\begingroup
\begin{equation}
\small
r_{lift} = \left\{\begin{array}{ll}
\min(2, 100 \times h ) & ,\ h \le 0.02 \\
15 - \min(5, 10 \times G({ori}_{t}, {ori}_{{target}})) \\
\quad - \min(5, 50 \times d({pos}_{t}, {pos}_{{target}})) & ,\ h > 0.02
\end{array}\right.
\end{equation}

\endgroup

We developed a piecewise lifting reward function \(r_{lift}\) to assist the dexterous hand in smoothly and accurately moving the object from the present position to the target position. The reward is linearly proportional to the lifting height \(h\) when \(h\le 0.02m\), and the maximum value is 2. This is intended to motivate the object to depart from its initial position and complete the fundamental lifting. The reward function transitions to the pose adjustment stage when \(h>0.02m\). It encourages the object to match the target pose by penalizing large orientation error and position error.

The complete reward is:

\begingroup
\small 
\begin{equation}
r =\alpha_{1} \times r_{approach} +CE \times \alpha_{2}\times  
 r_{grasp}+HT \times \alpha_{3}\times r_{lift}
\end{equation}
\endgroup
Where \(\alpha_{1},\alpha_{2},\alpha_{3}\) are the balance parameters,  set to 10, 10, and 20 respectively. They are kept constant during the training of all tasks.

\subsection{A.4 RL-Trainer}

Based on the task configuration obtained from RL-Configurator, we train a residual policy to output incremental action signals \(\Delta  a_t\) for fine-tuning the primary control signals \(A_{primary}\), enabling the robotic hand to adaptively optimize the contacts and grasping posture while maintaining the initial motion pattern. The observation state \(S_{t}\) includes: joint state \(q_{t}\), contact point coordinates \(C\) on the object surface, and the spatial positions of the fingertips \(P_{tips} \in R^{\mathbf{N\times3}}\), the spatial pose of the object \(o_{t}\), the target pose of the object \(\hat{o}\), the translational deviation \(E_{pos}\) between the current fingertip position and the corresponding grasping point, the rotation deviation \(D_{ori}\) between the current object pose and the target object pose and the normalized time step \(t \in R^{\mathbf{1}}\) , ranging  between 0 and 1. For translational deviations, the Euclidean distance is used for calculation \(d(x,y) =  \left\| \mathbf{x} - \mathbf{y} \right\|_{2}\), while the geodesic distance \(G\) is used for rotational deviation. We use the PPO algorithm for policy training, with a simple MLP network structure, a batch size of 64, a discount factor \(\gamma\) of 0.995, and a clipping coefficient \(\varepsilon \) of 0.3. All experiments are conducted in MuJoCo for training and testing, with a simulation frequency of 120 Hz, running on a computer with Ubuntu 20.04, an i7-7700 CPU, and a NVIDIA GeForce RTX 2080 Ti.

\section{Appendix B}

\subsection{B.1 Definition of the transfer success rate (TSR)}

To extract semantic-level action intentions from raw trajectory data, we determine transfer success rate based on the semantic similarity between the original and transferred trajectories. This method does not require the transfer trajectory to strictly reproduce each reference trajectory at spatial coordinate-level, but pays more attention to whether the action order is consistent. We extract the action intention sequence from the trajectory, through a sliding window, and evaluate the position and posture change characteristics in each window. Let the original trajectory be: \(O=\left\{\left(pos_{t}, ori_{t}\right)\right\}_{t=1}^{T}\), the window length set to 10, the sliding step set to 5, for each window, we extract the following action features: translation, tilt and rotation. According to the set threshold (the translation is 0.03 m, the tilt threshold is 15°, and the rotation threshold is 5°), each window is encoded as a semantic action intention: \(a_{i} \in \mathcal{A}=\{0,1,2, \ldots, K-1\}\), \(\mathcal{A}\) is a set of predefined semantic actions, 0: motionless, \textbf{1}: lift, \textbf{2}: fall, \textbf{3}: translation (horizontal), \textbf{4}: tilt (rotation along the x and y axes), \textbf{5}: rotation (rotation around the z axis). All identified semantic actions are concatenated according to the time sequence. At the same time, the low-amplitude noise segments are removed to improve semantic consistency and discrimination. Finally, a trajectory can be converted into an encoded action intention: \(S=\left\{a_{1}, a_{2}, \ldots, a_{N}\right\}\). Let the human action intention sequence be \(S^H=\left\{a^H_{1}, a^H_{2}, \ldots, a^H_{m}\right\}\), the dexterous hand action intention sequence be: \(S^R=\left\{a^R_{1}, a^R_{2}, \ldots, a^R_{n}\right\}\). To assess the semantic similarity between these two sequences, We define a distance matrix, where the distance between identical action labels is set to 0, and that between different labels is set to 1. Based on distance matrix we then compute the normalized Dynamic Time Warping (DTW) distance between the two sequences. If the normalized DTW distance is below a certain threshold (0.3 in our experiments), the transfer is considered successful. The transfer success rate TSR is calculated by the ratio of the number of tasks successfully transferred to the total number of tasks.

\subsection{B.2  Implementation of the D-Grasp algorithm}

D-Grasp  generates dynamically stable grasping sequences according to the predetermined grasping poses. The original implementation uses RaiSim. We reproduce its core logic in MuJoCo, like other compared methods, to reduce the impact of different simulators.

For  TCDM Task, we use our retargeting method to obtain the joint angle sequence that matches the human hand kinematics, and replay it in MuJoCo. We manually select the grasping frame, and record the dexterous hand joint position and wrist orientation and finger joint angle corresponding to the selected frame in the object coordinate system. For the contact vector, we directly use the human hand contact vector of the corresponding frame labeled by the GRAB dataset, and map it to the dexterous hand according to the corresponding relationship between the human hand and the dexterous hand. The above information is used as the predetermined static grasp reference. In addition, we use MuJoCo to obtain environmental information to construct the state space, including joint angles, joint angular velocities, hand-object contact force vectors, the 6D pose of the object and its angular velocity, the 6D pose of the wrist and its angular velocity, the displacement of the object relative to its initial position, the lifting height of the object, the difference between the current hand-object distance and the hand-object distance recorded in the static grasp reference, the angular distance between the current finger joint, wrist orientation and the rotation recorded in the static grasp reference, and the contact vector in the static grasp reference.

We implement the reward function of D-Grasp. The dexterous hand joint position is guided by calculating the weighted sum of squares of the Euclidean distance between the current position of the hand joint and the target position in the static grasp reference, the dexterous hand joint angle is guided by calculating the difference between the current hand joint angle and the target angle, the contact behavior of the dexterous hand is guided by calculating the proportion of the activated contact points to the target contact points and the force applied to the target contact points, and the high linear velocity and angular velocity of the hand and the object are penalized. The weighting parameters of the reward terms are set to be consistent with the original article.

The original D-Grasp experiment only focuses on simple relocation tasks. In order to compare with our method in various manipulation tasks, we divide the complete manipulation trajectory of the object and set multiple target references for D-Grasp. Specifically, a target reference is set every 50 time steps. For the reciprocating motion scene, we manually add a target reference at the reciprocating point.

\subsection{B.3 TCDM Task}

TCDM task set is used for performance comparison between different methods, selected from TCDM benchmark \cite{dasari2023learning}  totaling 40 manipulation tasks. The specific task list is presented in Table 6.

\begin{table}[t]
\centering
\begin{tabular}{|l|l|}
\hline
\multicolumn{2}{|c|}{\textbf{TCDM Task}}                   \\ \hline
airplane-fly1& knife-chop1        \\\hline
airplane-pass1& lightbulb-pass1    \\\hline
alarmclock-lift  & mouse-lift          \\\hline
alarmclock-see1  & mouse-use1          \\\hline 
banana-pass1     & mug-drink3          \\\hline
binoculars-pass  & piggybank-use1      \\\hline
cup-drink1       & scissors-use1       \\\hline
cup-pour1        & spheremedium-lift   \\\hline
duck-inspect1    & stapler-lift        \\\hline
duck-lift        & toothbrush-lift     \\\hline
elephant-pass1   & toothpaste-lift     \\\hline
eyeglasses-pass1 & toruslarge-inspect1 \\\hline
flashlight-lift  & train-play1         \\\hline
flashlight-on2   & watch-lift          \\\hline
flute-pass1      & waterbottle-lift    \\\hline
fryingpan-cook2  & waterbottle-shake1  \\\hline
hammer-use1      & wineglass-drink1    \\\hline
hand-inspect1    & wineglass-drink2    \\\hline
headphones-pass1 & wineglass-toast1   \\\hline
stamp-stamp1 & stanfordbunny-inspect1  
\\ \hline\end{tabular}
\caption{TCDM Task List}
\label{tab:my_table2}
\end{table}

\subsection{B.4 Additional results}

The quantitative results presented in Table 2 and Table 3 demonstrate PKDA's robustness against visual perception errors and its cross-hand transfer capability. Additional visualization results are presented below. Fig.9 showcases complete manipulation sequences successfully executed under significant perception errors. Fig.10 compares pre-grasp postures autonomously generated for Leap, Allegro, and Adroit. Fig.11 shows identical manipulation tasks executed across different hand platforms. 

Furthermore, we observed that the method's success rate decreases when manipulating objects of small sizes or slender shapes. As shown in Figure 12, for small-sized objects, their grasping points are typically more densely clustered. During the training process of PKDA, fingertip motion is guided by the grasping points and their neighboring regions. An excessively dense distribution of grasping points can easily lead to conflicts and confusion in fingertip exploration, thereby resulting in unreasonable grasp configurations. For slender or thin objects, due to their unique geometric structures, penetration is more likely to occur during hand-object interaction, further contributing to performance degradation.

In terms of transfer efficiency, we use the number of reinforcement learning iterations as the primary quantitative metric. To provide an intuitive demonstration of the transfer efficiency, we conducted experiments on 600 tasks in a hardware environment equipped with an Intel i7-7700 CPU and an NVIDIA GeForce RTX 2080 Ti. The transfer time for a single task ranged from 5 to 30 minutes, with an average of approximately 20 minutes. It is worth noting that a higher-performance CPU can support more parallel sampling of MuJoCo environments, thereby further reducing the transfer time.

\begin{figure}[t]
    \centering
    \includegraphics[width=1\linewidth]{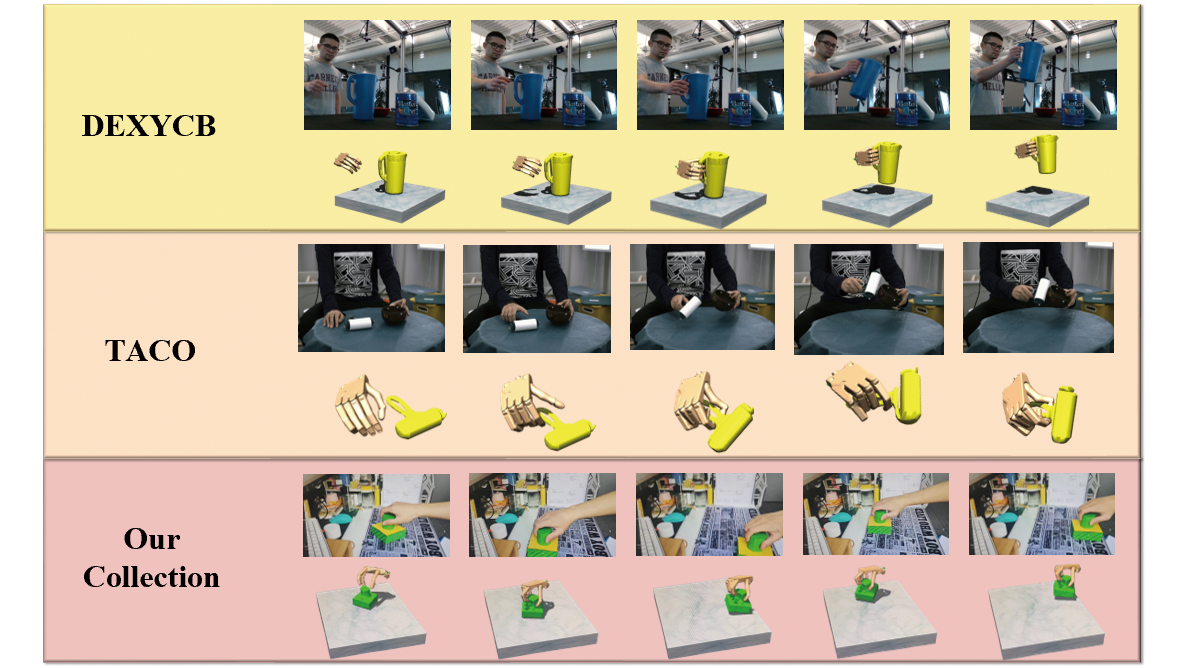}
    \caption{Transfer results of our PKDA framework in the presence of visual perception errors.
}
    \label{fig:enter-label9}
\end{figure}

\begin{figure*}[t]
    \centering
    \includegraphics[width=1\linewidth]{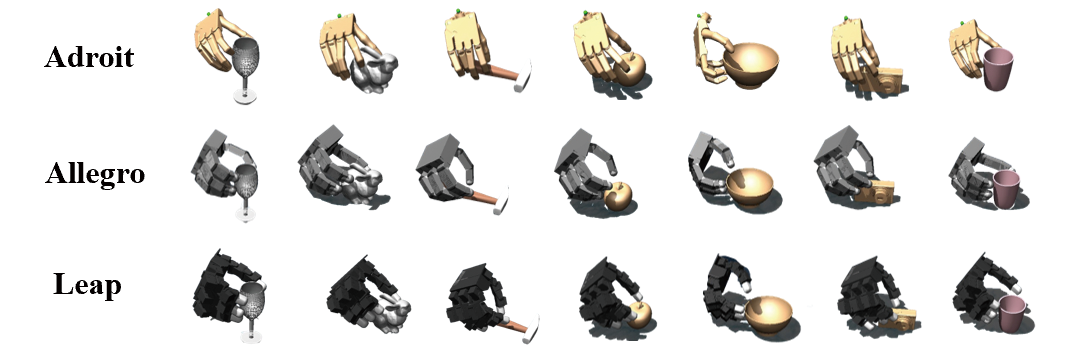}
    \caption{Pre-grasp across different dexterous hands}
    \label{fig:enter-label10}
\end{figure*}

\begin{figure*}[t]
    \centering
    \includegraphics[width=1\linewidth]{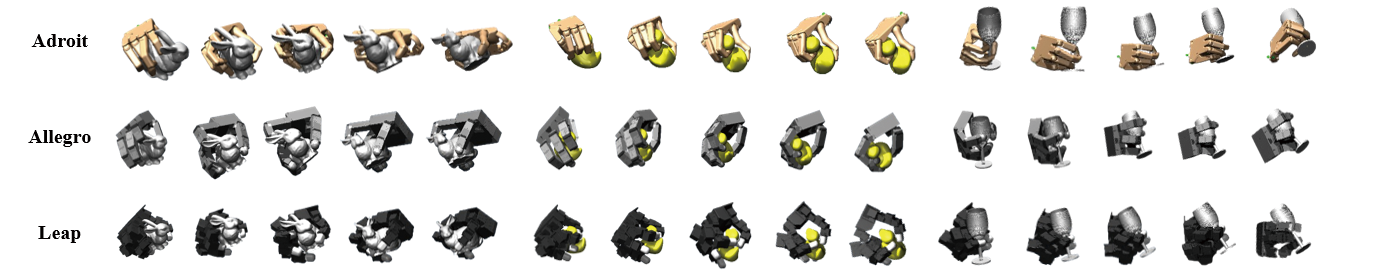}
    \caption{Comparative displays of different
dexterous hands executing identical manipulation tasks. }
    \label{fig:enter-label11}
\end{figure*}

\begin{figure}[t]
    \centering
    \includegraphics[width=1\linewidth]{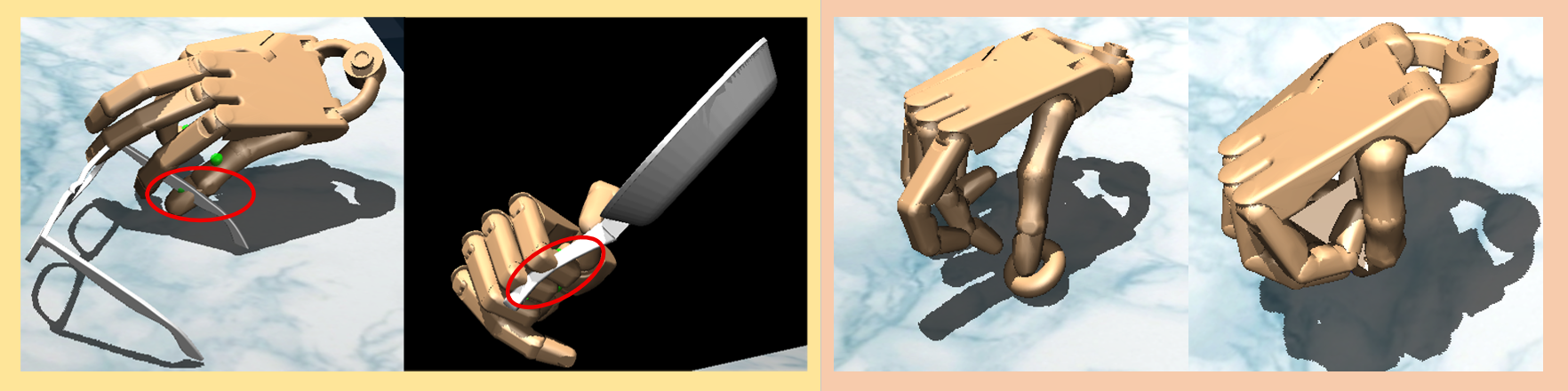}
    \caption{Interacting with thinner objects may lead to penetration (as shown in the left figure), while interacting with small-sized objects may result in unreasonable grasp configurations (as shown in the right figure).
}
    \label{fig:enter-label12}
\end{figure}

\section{Appendix C }

\paragraph{Task Settings}

In the real-world experiment, we designed three types of manipulation tasks with typical interactive semantics: shaking bottle, pouring with a cup and using a stamp .

\paragraph{Implementation process}

We first record videos of each manipulation task in a real-world environment. Subsequently, each video is processed with Hold\cite{fan2024hold} to extract the object model and the interaction trajectory between the human hand and the object. The resulting object model is imported into the simulator to construct the  simulation environment, and the extracted hand-object trajectory is input into the subsequent module of PKDA to generate the manipulation trajectory for Leap Hand. In the simulation, we adopted a floating dexterous hand model, in which the wrist movement was directly controlled by the virtual motor. In the real world, the Leap Hand is installed at the end of the UR10 robotic arm, and the wrist movement is physically constrained by the robotic arm. To achieve trajectory mapping from simulation to reality, we set specific marker points on the back of the Leap Hand in the simulation, and record the spatial trajectory of these marker points (representing the wrist pose) during the playback of the manipulation task, as well as the finger joint angles at each moment.

\paragraph{Open-loop control}

Referring to \cite{dasari2023learning}, we adopt an open-loop control mode to execute the entire manipulation sequences. Specifically, the motion trajectories of the wrist recorded in the simulation are taken as the target pose of the end of the UR10 robotic arm, and the recorded angles of the finger joints are taken as the target pose of the Leap Hand. Both are executed synchronously at the same frequency. In our observation, we find that this mode can achieve high-fidelity reproduction of the entire manipulation trajectories. 

\paragraph{Result}

Experiments show that driven by the simulation trajectory, Leap Hand can successfully reproduce the key interaction steps of typical manipulation tasks such as shaking, pouring and stamping. Without relying on real-world fine-tuning or additional supervision, our method demonstrates excellent action coherence and interaction stability in multiple manipulation tasks. 

\end{document}